\documentclass[runningheads]{llncs}
\usepackage{graphicx}

\usepackage{tikz}
\usepackage{comment}
\usepackage{amsmath,amssymb} 
\usepackage{color}
\usepackage{subfigure}
\usepackage{floatrow}
\newfloatcommand{capbtabbox}{table}[][\FBwidth]

\usepackage{blindtext}
\usepackage{hyperref}

\usepackage[accsupp]{axessibility}  
\usepackage{booktabs}
\usepackage{color, colortbl}
\definecolor{LightCyan}{rgb}{0.92,1,1}
\usepackage{multirow}

\begin{document}
\pagestyle{headings}
\mainmatter
\def\ECCVSubNumber{5443}  

\title{TDAM: Top-Down Attention Module for Contextually Guided Feature Selection in CNNs} 


\titlerunning{TDAM: Top-Down Attention Module for CNNs}
%
\author{Shantanu Jaiswal\inst{1} \and
Basura Fernando\inst{1,3} \and
Cheston Tan\inst{1, 2,3}}
\authorrunning{S. Jaiswal et al.}
%
\institute{Institute of High Performance Computing, A*STAR, Singapore 
\email{jaiswals@ihpc.a-star.edu.sg} \\ \and Institute for Infocomm Research, A*STAR, Singapore \\ \and Centre for Frontier AI Research, A*STAR, Singapore
}
\maketitle

\begin{abstract}
Attention modules for Convolutional Neural Networks (CNNs) are an effective method to enhance performance on multiple computer-vision tasks. While existing methods appropriately model channel-, spatial- and self-attention, they primarily operate in a feedforward bottom-up manner. Consequently, the attention mechanism strongly depends on the local information of a single input feature map and does not incorporate relatively semantically-richer contextual information available at higher layers that can specify “what and where to look” in lower-level feature maps through top-down information flow. 

\setlength\parindent{20pt} Accordingly, in this work, we propose a lightweight top-down attention module (TDAM) that iteratively generates a “visual searchlight” to perform channel and spatial modulation of its inputs and outputs more contextually-relevant feature maps at each computation step. Our experiments indicate that TDAM enhances the performance of CNNs across multiple object-recognition benchmarks and outperforms prominent attention modules while being more parameter and memory efficient. Further, TDAM-based models learn to “shift attention” by localizing individual objects or features at each computation step without any explicit supervision resulting in a 5\% improvement for ResNet50 on weakly-supervised object localization. Source code and models are publicly available at: \url{https://github.com/shantanuj/TDAM_Top_down_attention_module}.

\keywords{Object recognition; Visual attention mechanisms; Top-down feedback}
\end{abstract}

\section{Introduction}
\label{introd}
The design and incorporation of attention modules in deep CNNs has gained considerable recognition in computer vision due to their ability to enhance the representation power and performance of these networks in a task-agnostic manner. These modules typically formulate attention as a mechanism of feature modulation in outputs of traditional convolutional blocks by learning to intensify activations for deemed salient features and suppress activations for irrelevant ones. As a prominent method, Squeeze \& Excitation (SE) \cite{hu2018squeeze} introduces channel attention modelling of global-average-pooled (GAP) feature representations, which is then enhanced by CBAM \cite{woo2018cbam} through additional incorporation of spatial attention and utilization of both global-max-pooled (GMP) and GAP representations. Further, recent works \cite{qin2021fcanet,wang2020eca,gao2019global} identify how channel attention can be made more efficient and effective, while a different direction of work augments convolutional operations with self-attention and calibration methods \cite{Bello_2019_ICCV,liu2020improving} to learn more effective feature representations. 

However, conventional attention modules predominantly operate in a feedforward manner, i.e. they only utilize the output feature map of a convolutional block to both determine attention weights and perform attention modulation. As a result, the attentional mechanism is constrained to the representational capacity and local information of a single feature map input to the module. It does not incorporate relatively semantically-richer or task-specific contextual information available at higher layers while performing feature attention that can complement the initial bottom-up processing. This can be effectively facilitated by introducing top-down information flow between higher-level and lower-level feature representations within a convolutional block. Such feedback connections are also prevalent in the primate visual cortex \cite{kok2016selective} and recognized by neuroscientists as a key component in primate visual attention \cite{kreiman2020beyond,hochstein2002view}. Hence, in this work, we explore how top-down feedback computation can be effectively modelled to enable more contextually-guided feature activations across the CNN hierarchy. 

\begin{figure}[t]
\begin{center}
\includegraphics[width=0.98\linewidth]{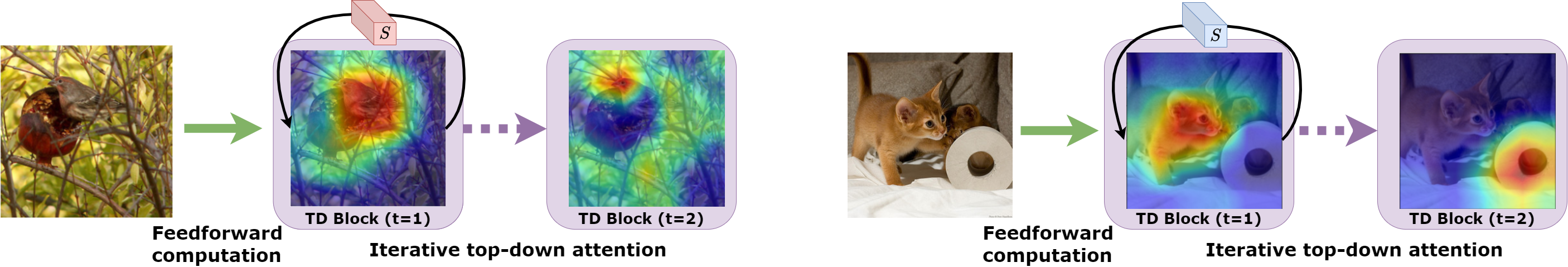}
\end{center}
\caption{Illustration of how iterative top-down feedback computation can help increase feature selectivity and thereby process individual salient features (left) or objects (right) at each computation step. Green arrows indicate feedforwarded inputs to TD blocks and red dotted arrows indicate top-down computation within TD block.} 
\label{fig:intro}
\end{figure}

\textbf{Top-down guided feature attention.}
A foundational formulation of top-down computation during visual processing was introduced by Crick in his ``searchlight hypothesis" \cite{crick1984function}, where he postulated the presence of an internal ``attentional searchlight" in the brain that operates by iteratively selecting lower-level neurons to ``co-fire" with semantically richer higher-level neurons, such that at any given instance, a sparse and strongly correlated set of selected lower and higher-level neurons fired together. 

Taking inspiration from his hypothesis, we illustrate how top-down feedback can guide feature attention in CNNs by taking an example of fine-grained bird classification. As shown in the first image input of figure \ref{fig:intro}, the initial feedforward computation provides a coarse feature activation which is sufficient to indicate that a ``bird-like" object is present, but is not precise to determine which exact ``bird". To guide feature attention, this top-level representation, which carries higher-level semantic information \cite{zeiler2014visualizing}, can be used to obtain an ``attentional searchlight'' \cite{crick1984function,treisman1980feature} that indicates which lower-level channels should be searched (in this case the particular beak type) and consequently does feedback computation to increase activation of the feature maps and spatial locations corresponding to selected lower-level channels. This results in a more precise lower-level representation which when consequently feedforwarded lends a more task-relevant top-level feature activation. The same mechanism for top-down guided feature attention can also enable the model to localize and process individual objects at each computation step as shown in second input of figure \ref{fig:intro}. 

While discussed for a single feedback step in the penultimate layer, the same operation can be done iteratively for multiple computation steps and at lower levels of the hierarchy. Note, in our discussion, we treat channel activation of intermediate feature maps to indicate presence of an individual feature and the corresponding 2D feature map of a channel to indicate the spatial location of a feature based on the findings in~\cite{zeiler2014visualizing}.

\textbf{Designing a top-down attention module with visual searchlights for feature attention.} Based on the above insights, we propose a novel top-down (TD) attention module that jointly models constituent higher-level and lower-level features to obtain a ``visual searchlight" that carries information on which lower-level features are of interest for subsequent computation. This searchlight then does attentional modulation of lower-level features by first performing channel attention through conventional channel scaling (``highlighting features of interest") and then performing spatial scaling (``intensifying spatial locations of highlighted features") by applying a spatial map obtained through its utilization in a single pointwise convolution. 

\textbf{Our proposed module is lightweight (in terms of parameters) and can be conveniently integrated} at multiple levels of the CNN hierarchy as a standard plug-in attention module and trained end-to-end with standard backpropagation. We discuss more details of our approach in section \ref{sec:approach}, and here briefly indicate two distinct advantages of the described operation of the visual searchlight -- (i) it enables task-specific and more informative activation of features at each computation step by localizing and processing individual features at each step (ii) it is more robust in performing spatial attention at changing input resolutions \cite{NEURIPS2019_d03a857a} in comparison to static convolutional kernel-based attention methods \cite{woo2018cbam,roy2018recalibrating}. 

\textbf{Contributions:} (i) We introduce a novel lightweight top-down attention module for CNNs by incorporating appropriate computational and neuroscience motivations in its design. (ii) We show the effectiveness of our module in enhancing performance of mainstream CNN models (ResNet, MobileNetV3 and ConvNeXt), outperforming state-of-the-art attention modules across multiple object recognition benchmarks, besides performing extensive ablation analysis to highlight key factors influencing the module's performance.  (iii) We demonstrate how our proposed module makes CNNs more robust to input resolution changes during inference and enables the emergent property of ``attention-shifting" through appropriate qualitative and quantitative analyses.

\section{Related work}
\label{sec:related_work}

\subsection{Attention modules for CNNs and feedforward attention mechanisms}

Initial work in the design of attention modules for CNNs includes the proposal of stacked attention modules for residual networks \cite{wang2017residual} and squeeze-and-excitation (SE) network's \cite{hu2018squeeze} channel attention formulation for feature aggregation and recalibration of GAP representations through fully connected layers. As extensions to SE, many works have focused on how to enhance the feature aggregation process by incorporating spatial and graph operations and more effective methods to estimate channel interactions than GAP. GE \cite{NEURIPS2018_dc363817} introduces a spatial gather-excite operator to augment channel attention, GSoP proposes second-order global pooling methods \cite{gao2019global}, C3 \cite{yang2019cross} incorporates graph convolutional networks for channel interactions and $A^2$-Nets \cite{NEURIPS2018_e1654211} incorporate second-order attentional pooling for long-range dependencies in image/video recognition. Notably, CBAM \cite{woo2018cbam} along with \cite{DBLP:conf/bmvc/ParkWLK18,roy2018recalibrating} demonstrate the advantage of incorporating spatial attention in conjunction to channel attention, while CBAM also indicates effectiveness of using GMP and GAP for feature aggregation. Further, AANets \cite{Bello_2019_ICCV} and SCNet\cite {liu2020improving} demonstrate how self-attention and self-calibration operations can augment standard convolutions, while GCNet \cite{cao2019gcnet} extends non-local neural networks to augment SE operations. More recently, prominent modules include ECA\cite{wang2020eca} which proposes one-dimensional convolutions to efficiently capture inter-channel interactions for channel attention and FCA\cite{qin2021fcanet} which proposes utilization of discrete cosine transform based frequency compression methods to effectively perform feature aggregation in SE in place of GAP. Additionally, modules such as DIANet\cite{huang2020dianet} and RLA\cite{zhao2021recurrence} propose to apply attention across layers, with the former applying a shared module across layers while the latter performs recurrent aggregation of features over layers.

In contrast to modelling attention for CNN architectures, recent works have studied how purely self-attention based Transformers \cite{vaswani2017attention} can be effectively applied in computer vision tasks. Starting from the first Vision Transformer \cite{dosovitskiy2020image}, refinements in both model design \cite{liu2021swin,wang2021not} and training strategies \cite{touvron2021training} have made Vision Transformers emerge as a strong class of vision backbones that utilize feedforward attention mechanisms for computer-vision tasks.

\subsection{Top-down feedback computation in CNNs}
Integrating top-down feedback computation in CNNs has in general been shown to improve performance on a variety of computer vision tasks. For instance, in neural image captioning and visual question answering, multiple methods employ variants of recurrent neural networks (RNNs) along with visual features from a CNN in an encoder-decoder setup \cite{chen2017sca,xu2015show,yang2016stacked,anderson2018bottom}. Similarly, RNNs have also been proposed to model visual attention for context-driven sequential computation in scene labelling \cite{pinheiro2014recurrent,byeon2015scene}, object recognition \cite{zhang2020putting,zamir2017feedback} and ``glimpse" based processing \cite{DBLP:journals/corr/BaMK14,NIPS2014_09c6c378} for multi-object classification. Finally, approaches also model top-down feedback to iteratively localize salient features \cite{fu2017look}, keypoints \cite{hu2016bottom} or objects \cite{cao2015look} and improve performance for fine-grained classification and object classification with cluttered inputs. While these approaches propose novel top-down feedback formulations, they are specifically formulated for target tasks and applied on existing ImageNet-1k pretrained backbone CNN models. In contrast, our top-down formulation serves a more general function of contextually-informed feature modulation of intermediate features across the CNN hierarchy and is integrated internally as a standard plug-in attention module trained with standard backpropagation. Consequently, we provide all together new ImageNet-1k pretrained backbones that we show to be effective on multiple object-recognition tasks.

\section{Top-down (TD) attention module}
\label{sec:approach}

\begin{figure}[t!]
\centering
\subfigure{\includegraphics[width=6.99cm]{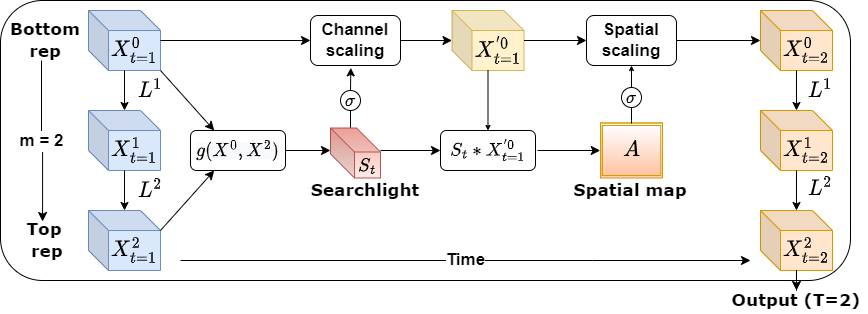}}
\subfigure{\raisebox{5mm}{\includegraphics[width=5cm, angle = 0]{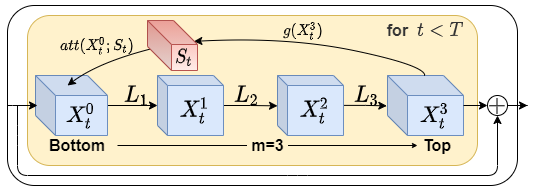}}}
\caption{\textbf{Left: Overview of our proposed top-down attention module} with ``joint" bottom and top attention (eq. \ref{eq: g_joint}). Given an input bottom feature map and feedforwarded top feature map, an ``attentional searchlight" $\mathcal{S}_t$ is inferred to perform channel and spatial attention of the existing input to obtain its next computational step representation, which is subsequently feedforwarded. This is repeated for T computational steps to obtain the final output $\mathbf{X^N_{t=T}}$ (N and T=2 in figure). `m' denotes the feedback distance between top and bottom representations. \textbf{Right: Integration of top-down (TD) module using ``top" attention (eq. \ref{eq: g_top}) in a ResNet \cite{he2016deep} bottleneck block.} As shown, TD operates before the residual connection for preset computational steps T and its output is added to the ResNet block input.}
\label{fig:Model_figure}
\end{figure}

We are given a convolutional block $\mathbf{B}$ comprising of $N$ convolutional layers, each denoted by $\mathbf{L_n}$ where $n \in \{1..N\}$,  that maps an input feature map $\mathbf{X^0}\in \mathbb{R}^{C_0 \times H_0 \times W_0}$  to an output feature map $\mathbf{X^N} \in \mathbb{R}^{C_N \times H_N \times W_N}$ through feedforward operation denoted by $\mathbf{L_N(L_{N-1}..(L_1(X^0)))}$. 
We denote the output of the top-down feedback operation in the block $\mathbf{B}$ for a given $t$ number of computational steps by $\mathbf{X^N_t}$ where $t \in \{1..T\}$.
Similarly, $\mathbf{X^0_t}$ is the input at computational step $t$.
We infer a 1D attentional searchlight $\mathbf{\mathcal{S}_t} \in \mathbb{R}^{C_0 \times 1 \times 1}$ that is used to sequentially perform channel and spatial attention of $\mathbf{X^0_t}$ to obtain the next computational step input $\mathbf{X^0_{t+1}}$, which is subsequently feedforwarded to obtain $\mathbf{X^N_{t+1}}$ as illustrated in figure \ref{fig:Model_figure}. 
The operations for each computational step $t$ can be summarized as follows:

\begin{align} \label{computation_summary}
\mathbf{X^N_t} &= \mathbf{L_N(L_{N-1}..(L_1(X^0_t)))} \\
\label{eqst2}
\mathbf{\mathcal{S}_t} &= \begin{cases}
                \mathbf{g(X^N_t, X^0_t)}  & \text{if joint attention} \\
                \mathbf{g(X^N_t)} & \text{if top attention}
                \end{cases}\\
\mathbf{X^0_{t+1}} &= \mathbf{att(X^0_t; \mathbf{\mathcal{S}_t})}
\end{align}
where $\mathbf{g(.)}$ is a learnable transformation and $\mathbf{att(.)}$ denotes channel and spatial attention. We provide more details for both below. The computation is repeated for $T$ computational steps and the final output of the block is $\mathbf{X^N_T}$.

\textbf{Obtaining the attentional searchlight.}\label{sec: att_searchlight}
In our proposed module, the attentional searchlight $\mathbf{\mathcal{S}_t}$ aims to specify which channels and spatial locations in a lower-level feature map should be emphasized for the next computational step and is derived from joint modelling of a higher-level feature map that captures a higher degree of semantic information  \cite{zeiler2014visualizing} and lower-level feature map that contains local feature information. 
Hence, we model the generation of $\mathbf{\mathcal{S}_t}$ as a joint learnable transformation of the higher-level feature map $\mathbf{X^N_t}$ and lower-level feature map $\mathbf{X^0_t}$. 
To perform the transformation, we first individually estimate channel activations by squeezing \cite{hu2018squeeze} the spatial dimensions of respective feature maps with an unparameterized pooling operation for both feature maps resulting in corresponding 1D channel vectors, for higher-level $\in \mathbb{R}^{C_N}$ and for lower-level $\in \mathbb{R}^{C_0}$. 
These vectors are then individually passed through distinct single hidden layer multi-layer perceptrons (MLP) and subsequently concatenated and passed through another single layer MLP to obtain a target 1D vector which we coin as \emph{Attention Searchlight} (i.e. $\mathbf{\mathcal{S}_t}$ given by equation~\ref{eqst2}). This computation is shown to be effective and efficient and also allows us to perform top-down feedback independent of the channel and spatial dimensions of top and bottom feature maps (i.e. $\mathbf{X^N_t}$ and $\mathbf{X^0_t}$ can be of different dimensions). The specific computation to obtain \emph{Attention Searchlight} is summarized as follows: 
\begin{align} \label{eq: g_joint}
\mathbf{\mathcal{S}_t} &=  \mathbf{g(X^N_t, X^0_t)} = \mathbf{W_s}(\texttt{ReLU}[\mathbf{W_{t}}(\mathbf{X^N_{t,p}}); \mathbf{W_{b}}(\mathbf{X^0_{t,p}  })]
\end{align}
where $\mathbf{W_t} \in \mathbb{R}^{C_N/r \times C_N}$, $\mathbf{W_b} \in \mathbb{R}^{C_0/r \times C_0}$ and $\mathbf{W_s} \in \mathbb{R}^{C_0 \times (C_N+C_0)/r}$ are weights of the MLP with ReLU activation applied after $\mathbf{W_t}$ and $\mathbf{W_b}$ , $r$ is a reduction ratio to reduce parameter complexity (we use 16), $\sigma$ is the sigmoid activation function and pooled representations are indicated by subscript $\mathbf{p}$, i.e., $\mathbf{X^N_{t,p}} = Pool(\mathbf{X^N_t})$ and we utilize spatial average-pooling for the $Pool()$ operator. Accordingly, we term the formulation in eq. \ref{eq: g_joint} as \textbf{joint attention} as it depends on both top and bottom feature-maps.

For cases where more parameter efficiency is desired (e.g. relatively large number of bottom channels), we model the generation of $\mathbf{\mathcal{S}_t}$ based on the hidden layer MLP representation of only the top feature map $\mathbf{X^N_t}$. We coin the below formulation as \textbf{top attention}:
\begin{align} \label{eq: g_top}
\mathbf{\mathcal{S}_t} &=  \mathbf{g(X^N_t)} = \mathbf{W_s}[\texttt{ReLU}(\mathbf{W_t}~(\mathbf{X^N_{t,p}})])
\end{align}
where $\mathbf{W_t} \in \mathbb{R}^{C_N/r \times C_N}$ and $\mathbf{W_s} \in \mathbb{R}^{C_0 \times C_N/r}$ are weights of the MLP with ReLU activation applied after $\mathbf{W_t}$, and $\sigma$ is the sigmoid activation function.

\textbf{Performing channel and spatial attention.}
We interpret the obtained attentional searchlight $\mathbf{\mathcal{S}}_t$ to first be a channel attention vector signifying which channels of $\mathbf{X^0_t}$ should be ``highlighted" for the next computation step. 
Consequently, as a first step, we scale the existing representation $\mathbf{X^0_t}$ through element-wise multiplication with $\mathbf{\mathcal{S}_t}$ (broadcasted spatially to match dimensions) to obtain its channel-scaled representation $\mathbf{X'^0_t}$. 
We then perform pointwise convolution of $\mathbf{\mathcal{S}_t}$ and $\mathbf{X'^0_t}$ with $\mathbf{\mathcal{S}_t}$ treated as a single 1x1 filter $\in \mathbb{R}^{1 \times 1 \times C_0}$ to obtain a 2D spatial map $\mathbf{\mathcal{A}} \in \mathbb{R}^{H_0 \times W_0}$ that specifies salient spatial locations in the scaled feature map $\mathbf{X'^0_t}$. 
Then, $\mathbf{X'^0_t}$ is scaled spatially through element-wise multiplication with the sigmoidal activation of $\mathbf{\mathcal{A}}$ (broadcasted channel-wise to match dimensions) to obtain the next computational-step input representation $\mathbf{X^0_{t+1}}$. We denote these set of operations as $\mathbf{att(X^0_t; \mathbf{\mathcal{S}_t})}$ and summarize it as:
\begin{align} \label{eq: att}
\mathbf{X'^0_{t}} &=  \mathbf{X^0_t} \otimes \sigma(\mathbf{\mathcal{S}_t}) \\ 
\mathbf{\mathcal{A}} &=  \mathbf{\mathcal{S}_t} \ast \mathbf{X'^0_t} \\ 
\mathbf{X^0_{t+1}} &=  \mathbf{X'^0_{t}} \otimes \sigma(\mathbf{\mathcal{A}}) 
\end{align}
where $\otimes$ denotes element-wise product and $\ast$ denotes pointwise convolution.

The intuition to perform pointwise convolution to obtain $\mathbf{\mathcal{A}}$ is that the channel weights in $\mathbf{\mathcal{S}_t}$ and increased activations for selected channels in $\mathbf{X'^0_t}$ ensure that only those spatial locations that correspond to selected channels are activated with convolution behaving as a spatial ``search" operation of selected lower-level features. A benefit of this formulation is that it can enable the model to be more robust to changes in input resolution that may occur during model inference and that can impact activation statistics of pooling layers \cite{NEURIPS2019_d03a857a} and spatial attention techniques that utilize fixed convolutional kernels \cite{woo2018cbam,roy2018recalibrating}.

\textbf{Integration in existing CNN models.}
As mentioned previously, our proposed module does not require the bottom input $\mathbf{X^0_t}$ to be of the same dimensions as the top output $\mathbf{X^N_t}$. Hence, it can be integrated into many CNN models as a standard attention module at multiple levels of the processing hierarchy and be trained end-to-end with standard backpropagation. 
Further, the formulation can be generalized to be a single block spanning the entire CNN model with the bottom input $\mathbf{X^0_t}$ = image input and $\mathbf{X^N_t}$ = pre-classifier feature map. 
However, we empirically find that having a large feedback-distance (number of feedforward convolutional layers) denoted by \textbf{`m'} between bottom representation $\mathbf{X^0_t}$ and the top representation $\mathbf{X^N_t}$ leads to unstable training and significantly worsen the performance. 
This is possibly due to a radical shift in input distributions over computational-steps for intermediate convolutional layers  (i.e. the layer receives two radically different inputs over computational-steps) due to changes accumulated over previous layer outputs that amplify with higher number of previous layers. 
Hence, for our experiments, we study \textbf{`m'} between 1 to 3 within a standard convolutional block, and specifically apply multiple instantiations of the module at deeper semantically-richer \cite{zeiler2014visualizing} levels of a CNN model. 
Further, we use unique batch normalization layers for each computation step to stabilize training as suggested in findings of \cite{liao2016bridging}. 
We denote our proposed top-down module which operates for \textbf{T} computational steps and over feedback-distance \textbf{M} with \textbf{TD (t=T, m=M)} where TD is further specified as top attention (eq. \ref{eq: g_top}) or joint attention (eq. \ref{eq: g_joint}).

An example integration of the module in a ResNet \cite{he2016deep} block is shown in figure \ref{fig:Model_figure}(right).

\section{Experimental results and discussion }
\label{sec:experiments}

\begin{table*}[t]
\small
\begin{center}
\resizebox{\columnwidth}{!}{
\begin{tabular}{|l|* {11}{c|}}
\hline
Method & BB. & Param. & FLOPs & \multicolumn{2}{|c|}{Mem.(Gb)} & \multicolumn{2}{|c|}{FPS/gpu} & \multicolumn{2}{|c|}{ImageNet-V2} & \multicolumn{2}{|c|}{ImageNet-V1}\\ 
\hline

\hline
- & - & - & - & Trn & Val & Trn & Val & Top1 & Top5 & Top1 & Top5\\ 

\hline
ResNet \cite{he2016deep} & \parbox[t]{1mm}{\multirow{4}{*}{\rotatebox[origin=c]{90}{Resnet50}}} & 25.56 M & 4.12 G & 29.5 & 16.1 & 704 & 2143 & 66.39 & 86.59 &  77.51 & 93.64\\

SE \cite{hu2018squeeze} & & 28.07 M & 4.13 G & 32.4 & 16.0 & 615 & 1911 &  66.92 & 86.88 & 78.03 & 93.88\\ 

CBAM \cite{woo2018cbam} &  & 28.07 M & 4.14 G & 37.6 & 20.7 & 420 & 1442 & 67.28 & 87.04 & 78.59 & 93.95\\ 

ECA \cite{wang2020eca}  & & 25.56 M & 4.13 G & 31.5 & 16.1 & 652 & 1989 & 66.72 & 86.95 & 78.11 & 93.85\\ 

FCA-TS \cite{qin2021fcanet} & & 28.07 M & 4.13 G & 32.4 & 16.3 & 590 & 1876 & 67.19 & 87.02 & 78.70 & 94.01\\ 

\rowcolor{LightCyan}
TDjoint (t=2, m=1) & & 27.65 M & 4.59 G & 31.9 & 16.2 & 601 & 1890 & 67.66 & 87.02 & \textbf{78.96} & 94.19\\

\rowcolor{LightCyan}
TDtop (t=2, m=1) & & 27.06 M & 4.59 G & 31.8 & 16.0 & 612 & 1905 & 67.21 & 86.98 & 78.82 & 93.98\\

\rowcolor{LightCyan}
TDtop (t=2, m=3) & & 27.66 M & 5.98 G & 35.3 & 16.3 & 498 & 1539 & \textbf{67.70} & \textbf{87.08} & 78.90 & \textbf{94.23}\\ 

\hline
ResNet \cite{he2016deep} & \parbox[t]{1mm}{\multirow{4}{*}{\rotatebox[origin=c]{90}{Resnet101}}} & 44.55 M & 7.85 G & 39.2 & 16.6 & 460 & 1376 & 69.64 & 89.09 & 80.36 & 95.31 \\ 

SE \cite{hu2018squeeze} &  & 49.29 M & 7.86 G & 45.5 & 16.9 & 368 & 1201 & 69.88 & 89.17 & 80.84 & 95.42\\ 

CBAM \cite{woo2018cbam} &  & 49.29 M & 7.88 G & 53.3 & 21.4 & 269 & 862 & 70.03 & 89.35 & 81.20 & 95.64\\ 

FCA-TS \cite{qin2021fcanet} & & 49.29 M & 7.86 G & 47.0 & 17.1 & 312 & 1164 & 70.12 & 89.42 & 81.15 & 95.59 \\

\rowcolor{LightCyan}
TDjoint (t=2, m=1) &  & 46.75 M & 8.37 G & 41.0 & 16.8 & 396 & 1237 & \textbf{70.56} & \textbf{89.44} & \textbf{81.62} & \textbf{95.76} \\
\rowcolor{LightCyan}
TDjoint (t=2, m=1, L4) &  & 45.94 M & 8.01 G & 40.3 & 16.8 & 413 & 1258 & 70.28 & 89.39 & 81.12 & 95.49 \\

\hline

\hline
\end{tabular}
}
\end{center}
\caption{Top1 \& Top5 single-crop classification accuracy (\%) of ResNet50 and ResNet101 integrated with our TD module in comparison to baselines on original ImageNet-V1 \cite{deng2009imagenet} and recent ImageNet-V2 \cite{recht2019imagenet} validation sets. All models are reproduced and trained with same experimental setup and selected on best ImageNet-V1 performance. Further backbones and baselines in supplemental for better readability.}\label{table: main_imnet} 
\vspace{-1em}
\end{table*}

We evaluate our proposed top-down (TD) attention module on the standard benchmarks: ImageNet-1k \cite{deng2009imagenet} for large-scale object classification and localization, CUB-200 \cite{WelinderEtal2010} and Stanford Dogs \cite{KhoslaYaoJayadevaprakashFeiFei_FGVC2011} for fine-grained classification and MS-COCO \cite{lin2014microsoft} for multi-label image classification. 

We perform experiments with three mainstream CNN model types -- ResNet \cite{he2016deep}, MobilenetV3 \cite{howard2019searching} and the recent ConvNeXt \cite{liu2022convnet}. We compare performance with the original models and prominent attention modules including Squeeze \& Excitation networks (SE) \cite{hu2018squeeze}, CBAM \cite{woo2018cbam}, ECA \cite{wang2020eca} and FCA \cite{qin2021fcanet}. To our knowledge, FCA is the most recent attention module shown to effectively enhance performance of multiple CNN variants.
For ResNet models, we apply our module at all blocks of layers 3 and 4 (with exception of ResNet101 wherein we either simply apply only at layer 4 or at layer 4 and 10 alternating blocks in layer 3 to preserve computational complexity in comparison to baselines). Similarly, for ConvNeXt, we apply at blocks of stages 3 and 4. For MobileNetV3 large, we apply our module at the final three layers, replacing the existing SE blocks in those layers.  We choose penultimate model layers as they generally constitute semantically-richer activations \cite{zeiler2014visualizing} where our described top-down attention mechanism can be most beneficial with marginal parameter overhead. For fair comparison, we reproduce all experiments in PyTorch \cite{NEURIPS2019_bdbca288} with the same training strategy used for all models. Training details and hyperparameter configurations for all experiments are provided in supplemental.

\begin{figure}[t!]
\begin{floatrow}
\capbtabbox{%
\scriptsize
\begin{tabular}{|l|c|c|c|c|} 
\hline
Model (RNet50) & \multicolumn{4}{|c|}{ImageNet-V1 Top1 Acc.} \\
\hline
- & Best & $224^2$ & $168^2$ & $448^2$   \\
\hline
ResNet \cite{he2016deep} & 78.24 & 77.51 & 74.53 & 75.64 \\

SE \cite{hu2018squeeze} & 78.75 & 78.03 & 75.52 & 76.78 \\

CBAM \cite{woo2018cbam} & 78.86 & 78.59 & 75.10 & 75.21 \\ 

ECA \cite{wang2020eca} & 78.80 & 78.11 & 75.46 & 76.85  \\

FCA-TS \cite{qin2021fcanet} & 79.02 & 78.70 & 75.74 & 76.99 \\

\rowcolor{LightCyan}
TDjoint(t2,m1) & \textbf{79.52} & \textbf{78.96} & 76.03 & 77.41 \\
\rowcolor{LightCyan}
TDtop(t2,m3) & 79.46 & 78.90 & \textbf{76.12} & \textbf{77.57}\\
\hline 
\end{tabular}
}{%
\caption{Performance of models with ResNet50 backbone on ImageNet-V1 single crop object classification at different testing resolutions (all models were trained on $224^2$ resolution). Most models obtain best accuracy at $280^2$ resolution as shown in fig. \ref{fig:res_plot}.}\label{table:res_table} 
}
\ffigbox{%
  \includegraphics[scale=0.45]{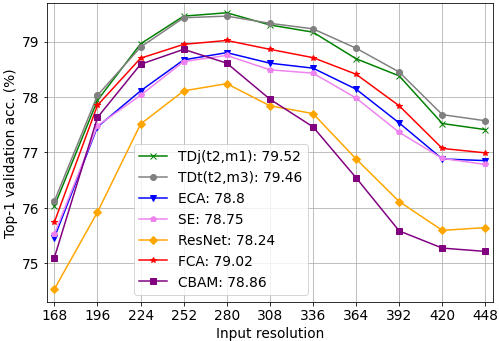}
}{%
  \caption{Performance of models (ResNet50 backbone) on ImageNet-V1 (ILSVRC-12 \cite{deng2009imagenet}) at different test resolutions with best accuracy reported in plot legend and table \ref{table:res_table}. TD models obtain better results at higher resolutions in comparison to baselines.} 
\label{fig:res_plot}
}

\end{floatrow}
\end{figure}

\subsection{Large-scale object classification (ImageNet-1k)}
\label{subsection: exp_imgnet}
We first perform experiments on large-scale object classification with the ImageNet-1k dataset and evaluate our module on ResNet variants (ResNet18, ResNet34, ResNet50 and ResNet101) and ConvNext-Tiny. For comprehensive evaluation, we consider two distinct validation sets -- the original ILSVRC-12 set comprising 50,000 images \cite{deng2009imagenet} and the recent more challenging ImageNet V2 \cite{recht2019imagenet} ``Matched-Frequency" set with 10,000 new images. We hereafter refer them as ImageNet-V1 and ImageNet-V2 respectively. We assess models based on their top1 and top5 single crop validation accuracy. Additionally, for models with our TD module, which output localized object predictions at each computation step (as shown in fig. \ref{fig:gradcam}), we only consider the most confident prediction during both training and evaluation with exception of a minority of images that comprise multiple objects, for which we consider only predictions with unique localization maps (having an IOU $<$ 0.5).

We summarize our experimental results for ResNet50 and ResNet101 in table \ref{table: main_imnet}, and compare the top configurations of our TD module with aforementioned baselines. 
We report results for other backbones in supplemental to not overload the reader and discuss ablations of our module in section \ref{sec:ablation}. 
We find that \textbf{our optimal formulation is TDjoint(t=2, m=1)}, which achieves a top1 accuracy of 78.96\%  on ImageNet-V1 and 67.66\% on ImageNet-V2 for ResNet50, improving performance of the original model by 1.5\% and 1.3\% respectively. For ResNet101, it achieves a 0.8\% higher top1 accuracy on ImageNet-V1 in comparison to SE when applied at both layers 3 and 4. Further, it outperforms feedforward attention modules on both validation sets for both ResNet50 and ResNet101 while having lesser parameters and comparable or higher training and inference speed in most cases (with exception of ECA). 
We find similar increments for TDtop(t=2, m=1), a lighter variant of our module utilizing only top attention, and TDtop(t=2, m=3), a computationally expensive variant with larger feedback distance.

Overall, our method outperforms all prior state-of-the-art attention modules including the recent FCA-TS~\cite{qin2021fcanet}, suggesting the effectiveness of the top-down attention mechanism of our module, which becomes more prominent for challenging tasks such as weakly supervised localization as shown later.


\textbf{Computation cost comparison:} Since our modules perform iterative top-down computation, they have a higher associated number of FLOPs than existing feedforward modules. Further, the number of FLOPs grows as the feedback distance (`m') increases -- growing by 20\% in case of ResNet50 from m=1 to m=3. 
However, we find that this limitation can be managed in practice during both training and inference in comparison to baseline modules since our models (specifically with m=1) require lesser parameters and memory operations, and hence have comparable or higher FPS (speed). 
As shown in table \ref{table: main_imnet}, TD models with m=1 have higher FPS than both CBAM and FCA for ResNet50, and higher FPS than SE as well for ResNet101. 

\textbf{Evaluation of models at different input resolutions.} The activation statistics of the global pooling in CNNs have been shown to be strongly impacted by changes in input resolutions \cite{NEURIPS2019_d03a857a}, thereby making performance of CNNs susceptible to variations in input resolution. Hence, in this experiment, we study whether attention modules including TD can enhance robustness of CNNs by evaluating ResNet50 models at different resolutions for ImageNet-V1. We plot results for all models in fig. \ref{fig:res_plot} for testing resolutions from $168$x$168$ to $448$x$448$ with increments of $28$ (further lower resolutions provided in supplemental). Additionally, in table \ref{table:res_table}, we indicate each model's performance at lowest resolution of $168$x$168$ (denoted by $168^2$), performance at highest resolution of $448$x$448$, best testing performance and original $224$x$224$ performance. 

We find that: (i) TD-models largely prevent performance from dropping drastically at higher resolutions, particularly obtaining a 2\% better performance than the original ResNet model at $448$x$448$. (ii) CBAM, which utilizes a fixed convolutional kernel to model spatial attention, has a significant drop in performance at higher resolutions. (iii) Other attention modules have appreciable robustness in comparison to original ResNet, but have lesser benefits at higher resolutions in comparison to TD-models, with the best performing module (FCA) having a 0.5\% lesser performance than TD-models at best resolution setting.

\subsection{Attention visualization and weakly-supervised object localization}
\label{subsection: exp}
\begin{figure*}[t]
\includegraphics[width=0.99\columnwidth]{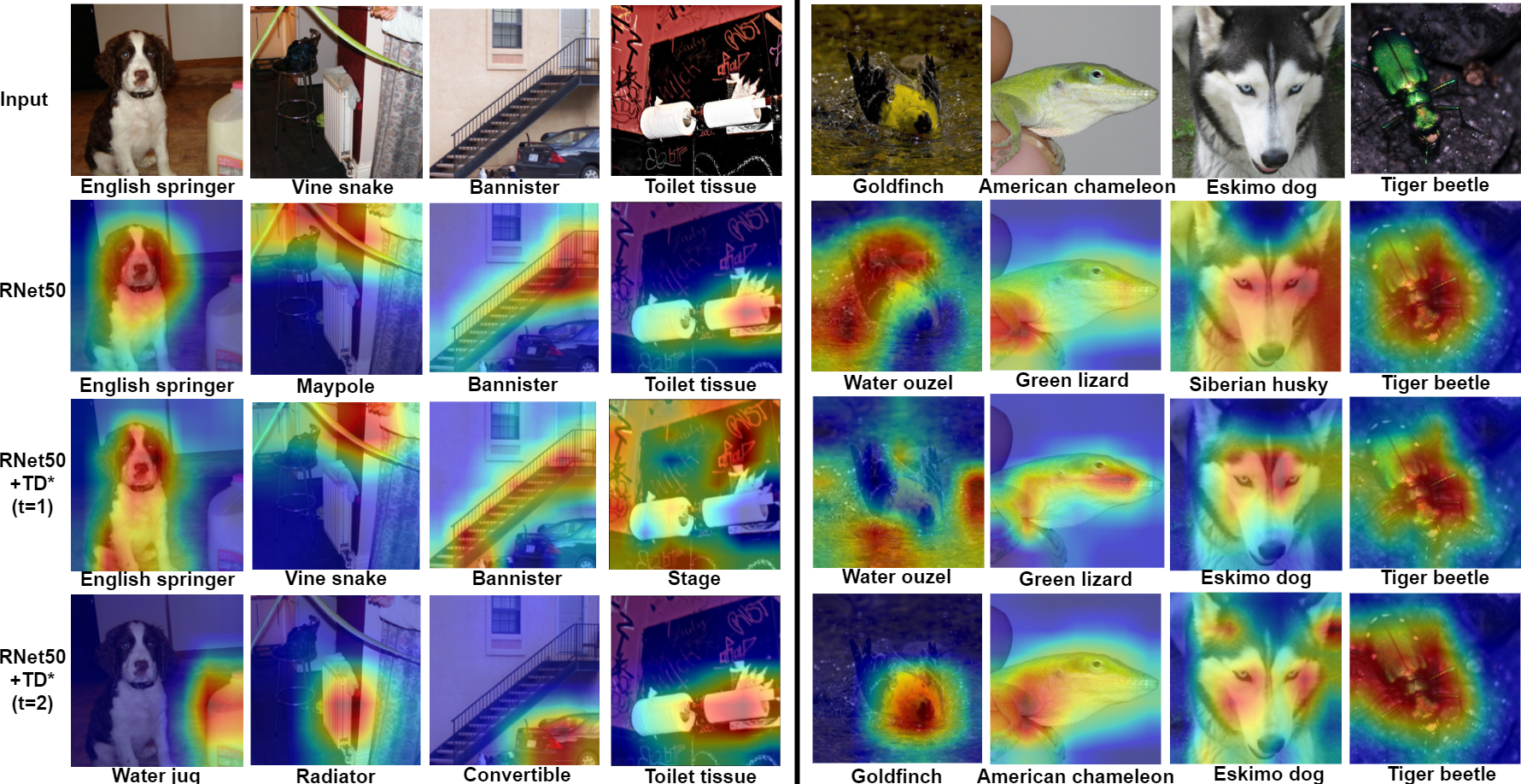}
\caption{\textbf{Representative examples of ``attention shifting" over computational steps of our model based on Grad-CAM analysis}. In the first 4 examples, the TD model iteratively attends to distinct objects and has a more selective and complete feature activation at each computation step compared to original ResNet50. In the next 4 examples, it iteratively attends to relevant features for better discrimination of finer classes. See  supplemental for more types and code to generate arbitrary examples.} 
\label{fig:gradcam}
\end{figure*}

To better understand the workings of our proposed TD module, we utilize Grad-CAM maps \cite{selvaraju2017grad} to visualize the model's attention at each computation step for images drawn from the aforementioned validation sets. As shown in figure \ref{fig:gradcam}, we find that the model implicitly learns to shift its attention over computation steps. We conjecture that this imparts the model two important capabilities -- one, choosing which object to ``focus" on at each computational step when multiple objects (known to the model) are present and two, which feature to ``focus" on at a given computational step when a more ambiguous or ``difficult" object is present as the input. As an example of the first case, consider the second image input in fig. \ref{fig:gradcam}, wherein the model at its first computational step accurately locates a `vine snake' to be present in the scene (which the original model gets incorrect) and then shifts its attention to the radiator. In contrast, in input five (of same fig. \ref{fig:gradcam}), the model iteratively attends to different features to make a more informed prediction. At its first computational step, it identifies water as a primary feature and has an initial prediction of a `water-ouzel', but in the second computational step, it shifts its attention to the head of the bird and consequently predicts the correct category -- `goldfinch'. In comparison, the original Resnet50's prediction (`water-ouzel') is based on a less selective and conjoined feature map of the bird and water. 

To quantitatively assess the model's attention capability and resulting enhancement in feature selectivity, we evaluate it on weakly supervised ImageNet-1k localization challenge \cite{deng2009imagenet}, which requires models to provide bounding boxes in addition to classification labels. For all models, we follow the same strategy as \cite{selvaraju2017grad} to generate bounding boxes for output predicted classes, and report top-1 and top-5 localization accuracy on ImageNet-V1 in table \ref{weakly-supervised}. We find that both top performing TD configurations for object classification improve performance of ResNet50, with TDtop(t=2,m=3) notably increasing accuracy by 5\%, and being 3\% over the best baseline model -- CBAM. Interestingly, channel attention methods of SE, ECA and FCA obtain worse performance than original ResNet50, suggesting strong importance of spatial attention in localization tasks. 
The resulting improvement also highlights the importance of the top-down searchlight-driven feedback mechanism introduced in this paper for obtaining a better attention module.

\begin{figure}
\begin{floatrow}
\capbtabbox{%
\scriptsize
\begin{tabular}{|l|c|c|} 
\hline
Model & \multicolumn{2}{|c|}{ImageNet(V1)} \\
\hline
- & Top1 & Top5   \\
\hline
RNet50 & 57.04 & 68.67  \\

RNet101 & 58.54 & 69.86  \\

RNet50 + SE & 56.62 & 67.88 \\ 

RNet50 + CBAM & 58.91 & 70.54  \\

RNet50 + ECA & 56.94 & 68.38  \\

RNet50 + FCA-TS & 56.88 & 67.86  \\
\rowcolor{LightCyan}
RNet50 + TDjoint(t=2,m=1) & 61.55 & 72.10 \\
\rowcolor{LightCyan}
RNet50 + TDtop(t=2,m=3) & \textbf{61.97} & \textbf{72.37}\\

\hline 
\end{tabular}
}{%
  \caption{Weakly supervised object localization accuracy (\%) on ImageNet (V1). TD models and CBAM that incorporate spatial attention increase performance of ResNet50 while purely channel attention methods reduce performance.}\label{weakly-supervised}
}
\capbtabbox{%
\scriptsize
\begin{tabular}{|l|c|c|c|c|} 
\hline
Model (ResNet50) & CUB & Dogs & \multicolumn{2}{|c|}{MS-COCO} \\ 
\hline
- & Top1 & Top1 & mAP & F1-O \\ 
\hline\hline
ResNet & 88.26 & 85.97 & 77.58 & 75.45 \\ 
SE & 88.89 & 86.55 & 78.21 & 76.37\\ 
CBAM & 89.37 & 86.98 & 79.17 & 77.15\\ 
FCA-TS & 88.94 & 86.76 & 79.05 & 77.08\\ 
\rowcolor{LightCyan}
TDjoint(t=2,m=1) & 89.61 & 87.08 & \textbf{79.61} & \textbf{77.71}\\
\rowcolor{LightCyan}
TDtop(t=2,m=3) & \textbf{89.75} & \textbf{87.30} & 79.56 & 77.62 \\ 
\hline
\end{tabular}
}{%
  \caption{Performance of models as backbones for fine-classification (val. acc. \% for CUB and Stanford Dogs) and multi-label classification (val. mAP and overall F1 for MS-COCO). TD-based ResNet50 backbones outperform baselines in both tasks.}\label{table: fine_class}
}
\end{floatrow}
\end{figure}

\subsection{Fine-grained and multi-label classification}
To demonstrate the general applicability of our TD module across different tasks and assess its capability as a robust feature extractor, we evaluate its performance as a backbone for existing state-of-the-art methods for fine-grained classification and multi-label object classification. We use the ``Weakly Supervised Data Augmentation" method \cite{hu2019see} for fine-grained classification and ``Asymmetric loss" method \cite{ridnik2021asymmetric} for multi-label classification, and simply replace the model backbone in both methods with our pretrained ImageNet-1k models. For fine-grained classification, we consider the Caltech-birds (CUB) and Stanford Dogs (Dogs) datasets and assess models on top-1 validation accuracy. For multi-label classification, we use MS-COCO and assess models on mAP and overall F1. 

As shown in table \ref{table: fine_class}, using TD models (denoted as TDj for TDjoint and TDt for TDtop) as a backbone leads to a notable improvement in all three tasks compared to a baseline ResNet50, specifically achieving 1.5\% increment on CUB-200 and 2 points better mAP on MS-COCO. The relative improvement over CBAM, which performs purely feedforward channel and spatial attention, suggests the benefits of feedback-driven channel and spatial attention in enabling iterative task-specific refinement of constituent feature-maps within the backbone. 

\subsection{Ablative analysis of feedback computation}\label{sec:ablation}
As indicated in Sec. \ref{sec:approach}, our proposed TD module has four primary factors -- (i) choice of attention ``TDjoint" or ``TDtop" (ii) feedback distance `m', (iii) feedback channel and spatial attention technique, and (iv) feedback steps `t'. Accordingly, we assess the impact of each factor by evaluating resulting module configurations on ImageNet classification. To evaluate factors (i), (ii) and (iii), we utilize ResNet50. For factor (iv), we utilize relatively shallower models of MobileNetV3-large and ResNet18 and evaluate on a hierarchically reduced subset of ImageNet with 200 classes due to the high computation cost and training time associated with models with more than 3 feedback steps. 

\begin{figure}[t]
\begin{center}
\includegraphics[width=1\linewidth]{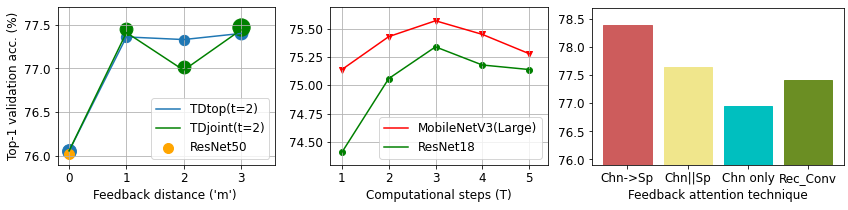}
\end{center}
\caption{\textbf{Ablative analysis of our TD module.} \textbf{Left}: choice of attention operation (TDjoint or TDtop) and feedback distance `m' (with ResNet50 on ImageNet-1k). \textbf{Center}: number of feedback steps `t' (for MobileNetV3 large and ResNet18 on a hierarchically reduced subset of ImageNet with 200 classes). 
\textbf{Right}: Impact of feedback attention technique (with ResNet50) where Chn$\rightarrow$ Sp denotes spatial attention performed after channel, Chn$||$Sp denotes independent parallel channel and spatial attention, and Conv\_Map denotes a feedback convolution. Numerical reports in supplemental.} 
\label{fig:td_ablation}
\vspace{-1em}
\end{figure}

\textbf{Impact of feedback distance `m' and ``joint" vs ``top" attention} In the left plot of fig.\ref{fig:td_ablation}, we see that having a feedback distance of at least 1 improves the performance of the module, i.e. the module requires distinct top and bottom feature maps, and applying top-down attention on the same single feature map as done in existing attention modules provides negligible performance benefit over the baseline ResNet50 while introducing high number of parameters. Next, the performance is most improved at m=1 and m=3 for both TDjoint and TDtop. However, note this does not indicate that m=2 is an inferior option in general, as in the case of ResNet50, the bottom representation at m=2 is the bottleneck block input, which may not sufficiently benefit from attentional modulation. Finally, TDjoint(m=3) and TDjoint(m=1) are the top-2 performing modules configurations indicating the enhanced representation capacity offered in joint modelling of top and bottom feature maps. However, while TDjoint(m=3) has the highest performance, it has 8\% higher parameters than TDjoint(m=1) and TDtop(m=3). Hence, other variants are more preferable, and in particular, \textbf{we recognize TDjoint (t=2, m=1) as our primary TD attention module}.

\textbf{Impact of feedback steps `t'.} As shown in center plot of fig. \ref{fig:td_ablation}, the performance peaks at 2 feedback steps, and thereafter declines, but still retains higher performance than both a purely feedforward (t=1) model and single feedback (t=2) model. We conjecture that this decline at higher computation steps may be a result of possibly reduced value of gradients while training of the model (akin to ``vanishing gradients"), leading to a less effective attentional searchlight at each feedback step. Note that in case of MobileNet, models with TD exhibit higher performance than the original model while having 15\% lesser parameters and smaller model size, suggesting TD attention is useful for low-end devices.

\textbf{Impact of feedback attention technique.} To study the individual contributions of channel and spatial attention, as well as the applied order of attention operations, we consider six variants of the feedback operation in our TD module -- (i) channel then spatial attention (as described in sec. \ref{sec:approach}), (ii) spatial then channel attention, (iii) channel and spatial attention performed independently and parallely, (iv) only channel attention, (v) only spatial attention, and (vi) use of a recurrently-applied convolutional layer to map output feature maps to next computation step input feature maps (instead of explicitly performing attention). As shown in the right plot of fig. \ref{fig:td_ablation}, we find that performing channel and spatial attention independently has lesser performance than our searchlight's intended operation of first performing channel attention and then spatial attention. Additionally, only doing channel attention led to a noticeable drop in performance indicating the contribution of spatial attention in the searchlight's operation. Similarly, iteratively applying a convolutional layer on the output feature map to obtain next computation step inputs had lesser performance indicating the benefit of the iterative top-down attention method utilized in our work. For only spatial attention and spatial followed by channel attention, we found the network did not converge while training and had significantly worse performance.
Apart from the above-mentioned primary factors of our module, we also quantitatively study how our module impacts selectivity of channels in its output feature maps, and report other ablations 
in the supplemental.

\section{Conclusions}
\label{sec:conclusions}
We introduced a lightweight module for CNNs that iteratively generates a ``visual searchlight" to perform top-down channel and spatial attention of its constituent representations and outputs more selective feature activations. We performed extensive experiments with mainstream CNNs and showed that integrating our module outperforms baseline attention modules on large-scale object classification, fine-grained and multi-label classification. Further, we demonstrated the effectiveness of TD-models in increasing robustness to changes in image resolutions during inference and also illustrated the emergent ``attention shifting" behaviour and quantitatively assessed it on weakly supervised object localization, finding that it outputs significantly more precise localization maps. This can be especially beneficial for applications with varying input resolutions or requiring fine-resolution processing such as ego-centric action anticipation.\\

\noindent
\small{
\textbf{Acknowledgment}
This research/project is supported in part by the National Research Foundation, Singapore under its AI Singapore Program (Award Number: AISG-RP-2019-010). This research is also supported by funding allocation to C.T. and B.F. by the Agency for Science, Technology and Research (A*STAR) under its SERC Central Research Fund (CRF), as well as its Centre for Frontier AI Research (CFAR).} 

\clearpage
%
%
\bibliographystyle{splncs04}
\bibliography{2_bib}

\begin{thebibliography}{10}
\providecommand{\url}[1]{\texttt{#1}}
\providecommand{\urlprefix}{URL }
\providecommand{\doi}[1]{https://doi.org/#1}

\bibitem{anderson2018bottom}
Anderson, P., He, X., Buehler, C., Teney, D., Johnson, M., Gould, S., Zhang,
  L.: Bottom-up and top-down attention for image captioning and visual question
  answering. In: Proceedings of the IEEE conference on computer vision and
  pattern recognition. pp. 6077--6086 (2018)

\bibitem{DBLP:journals/corr/BaMK14}
Ba, J., Mnih, V., Kavukcuoglu, K.: Multiple object recognition with visual
  attention. In: ICLR (Poster) (2015), \url{http://arxiv.org/abs/1412.7755}

\bibitem{Bello_2019_ICCV}
Bello, I., Zoph, B., Vaswani, A., Shlens, J., Le, Q.V.: Attention augmented
  convolutional networks. In: Proceedings of the IEEE/CVF International
  Conference on Computer Vision (ICCV) (October 2019)

\bibitem{byeon2015scene}
Byeon, W., Breuel, T.M., Raue, F., Liwicki, M.: Scene labeling with lstm
  recurrent neural networks. In: Proceedings of the IEEE Conference on Computer
  Vision and Pattern Recognition. pp. 3547--3555 (2015)

\bibitem{cao2015look}
Cao, C., Liu, X., Yang, Y., Yu, Y., Wang, J., Wang, Z., Huang, Y., Wang, L.,
  Huang, C., Xu, W., et~al.: Look and think twice: Capturing top-down visual
  attention with feedback convolutional neural networks. In: Proceedings of the
  IEEE international conference on computer vision. pp. 2956--2964 (2015)

\bibitem{cao2019gcnet}
Cao, Y., Xu, J., Lin, S., Wei, F., Hu, H.: Gcnet: Non-local networks meet
  squeeze-excitation networks and beyond. In: Proceedings of the IEEE/CVF
  International Conference on Computer Vision Workshops. pp.~0--0 (2019)

\bibitem{chen2017sca}
Chen, L., Zhang, H., Xiao, J., Nie, L., Shao, J., Liu, W., Chua, T.S.: Sca-cnn:
  Spatial and channel-wise attention in convolutional networks for image
  captioning. In: Proceedings of the IEEE conference on computer vision and
  pattern recognition. pp. 5659--5667 (2017)

\bibitem{NEURIPS2018_e1654211}
Chen, Y., Kalantidis, Y., Li, J., Yan, S., Feng, J.: A\^{}2-nets: Double
  attention networks. In: Bengio, S., Wallach, H., Larochelle, H., Grauman, K.,
  Cesa-Bianchi, N., Garnett, R. (eds.) Advances in Neural Information
  Processing Systems. vol.~31. Curran Associates, Inc. (2018),
  \url{https://proceedings.neurips.cc/paper/2018/file/e165421110ba03099a1c0393373c5b43-Paper.pdf}

\bibitem{crick1984function}
Crick, F.: Function of the thalamic reticular complex: the searchlight
  hypothesis. Proceedings of the National Academy of Sciences  \textbf{81}(14),
   4586--4590 (1984)

\bibitem{deng2009imagenet}
Deng, J., Dong, W., Socher, R., Li, L.J., Li, K., Fei-Fei, L.: Imagenet: A
  large-scale hierarchical image database. In: 2009 IEEE conference on computer
  vision and pattern recognition. pp. 248--255. Ieee (2009)

\bibitem{dosovitskiy2020image}
Dosovitskiy, A., Beyer, L., Kolesnikov, A., Weissenborn, D., Zhai, X.,
  Unterthiner, T., Dehghani, M., Minderer, M., Heigold, G., Gelly, S., et~al.:
  An image is worth 16x16 words: Transformers for image recognition at scale.
  arXiv preprint arXiv:2010.11929  (2020)

\bibitem{fu2017look}
Fu, J., Zheng, H., Mei, T.: Look closer to see better: Recurrent attention
  convolutional neural network for fine-grained image recognition. In:
  Proceedings of the IEEE conference on computer vision and pattern
  recognition. pp. 4438--4446 (2017)

\bibitem{gao2019global}
Gao, Z., Xie, J., Wang, Q., Li, P.: Global second-order pooling convolutional
  networks. In: Proceedings of the IEEE/CVF Conference on Computer Vision and
  Pattern Recognition. pp. 3024--3033 (2019)

\bibitem{he2016deep}
He, K., Zhang, X., Ren, S., Sun, J.: Deep residual learning for image
  recognition. In: Proceedings of the IEEE conference on computer vision and
  pattern recognition. pp. 770--778 (2016)

\bibitem{hochstein2002view}
Hochstein, S., Ahissar, M.: View from the top: Hierarchies and reverse
  hierarchies in the visual system. Neuron  \textbf{36}(5),  791--804 (2002)

\bibitem{howard2019searching}
Howard, A., Sandler, M., Chu, G., Chen, L.C., Chen, B., Tan, M., Wang, W., Zhu,
  Y., Pang, R., Vasudevan, V., et~al.: Searching for mobilenetv3. In:
  Proceedings of the IEEE/CVF International Conference on Computer Vision. pp.
  1314--1324 (2019)

\bibitem{NEURIPS2018_dc363817}
Hu, J., Shen, L., Albanie, S., Sun, G., Vedaldi, A.: Gather-excite: Exploiting
  feature context in convolutional neural networks. In: Bengio, S., Wallach,
  H., Larochelle, H., Grauman, K., Cesa-Bianchi, N., Garnett, R. (eds.)
  Advances in Neural Information Processing Systems. vol.~31. Curran
  Associates, Inc. (2018),
  \url{https://proceedings.neurips.cc/paper/2018/file/dc363817786ff182b7bc59565d864523-Paper.pdf}

\bibitem{hu2018squeeze}
Hu, J., Shen, L., Sun, G.: Squeeze-and-excitation networks. In: Proceedings of
  the IEEE conference on computer vision and pattern recognition. pp.
  7132--7141 (2018)

\bibitem{hu2016bottom}
Hu, P., Ramanan, D.: Bottom-up and top-down reasoning with hierarchical
  rectified gaussians. In: Proceedings of the IEEE Conference on Computer
  Vision and Pattern Recognition. pp. 5600--5609 (2016)

\bibitem{hu2019see}
Hu, T., Qi, H., Huang, Q., Lu, Y.: See better before looking closer: Weakly
  supervised data augmentation network for fine-grained visual classification.
  arXiv preprint arXiv:1901.09891  (2019)

\bibitem{huang2020dianet}
Huang, Z., Liang, S., Liang, M., Yang, H.: Dianet: Dense-and-implicit attention
  network. In: Proceedings of the AAAI Conference on Artificial Intelligence.
  vol.~34, pp. 4206--4214 (2020)

\bibitem{KhoslaYaoJayadevaprakashFeiFei_FGVC2011}
Khosla, A., Jayadevaprakash, N., Yao, B., Fei-Fei, L.: Novel dataset for
  fine-grained image categorization. In: First Workshop on Fine-Grained Visual
  Categorization, IEEE Conference on Computer Vision and Pattern Recognition.
  Colorado Springs, CO (June 2011)

\bibitem{kok2016selective}
Kok, P., Bains, L.J., van Mourik, T., Norris, D.G., de~Lange, F.P.: Selective
  activation of the deep layers of the human primary visual cortex by top-down
  feedback. Current Biology  \textbf{26}(3),  371--376 (2016)

\bibitem{kreiman2020beyond}
Kreiman, G., Serre, T.: Beyond the feedforward sweep: feedback computations in
  the visual cortex. Annals of the New York Academy of Sciences
  \textbf{1464}(1),  222--241 (2020)

\bibitem{liao2016bridging}
Liao, Q., Poggio, T.: Bridging the gaps between residual learning, recurrent
  neural networks and visual cortex. arXiv preprint arXiv:1604.03640  (2016)

\bibitem{lin2014microsoft}
Lin, T.Y., Maire, M., Belongie, S., Hays, J., Perona, P., Ramanan, D.,
  Doll{\'a}r, P., Zitnick, C.L.: Microsoft coco: Common objects in context. In:
  European conference on computer vision. pp. 740--755. Springer (2014)

\bibitem{liu2020improving}
Liu, J.J., Hou, Q., Cheng, M.M., Wang, C., Feng, J.: Improving convolutional
  networks with self-calibrated convolutions. In: Proceedings of the IEEE/CVF
  Conference on Computer Vision and Pattern Recognition. pp. 10096--10105
  (2020)

\bibitem{liu2021swin}
Liu, Z., Lin, Y., Cao, Y., Hu, H., Wei, Y., Zhang, Z., Lin, S., Guo, B.: Swin
  transformer: Hierarchical vision transformer using shifted windows. In:
  Proceedings of the IEEE/CVF International Conference on Computer Vision. pp.
  10012--10022 (2021)

\bibitem{liu2022convnet}
Liu, Z., Mao, H., Wu, C.Y., Feichtenhofer, C., Darrell, T., Xie, S.: A convnet
  for the 2020s. arXiv preprint arXiv:2201.03545  (2022)

\bibitem{NIPS2014_09c6c378}
Mnih, V., Heess, N., Graves, A., kavukcuoglu, k.: Recurrent models of visual
  attention. In: Ghahramani, Z., Welling, M., Cortes, C., Lawrence, N.,
  Weinberger, K.Q. (eds.) Advances in Neural Information Processing Systems.
  vol.~27. Curran Associates, Inc. (2014),
  \url{https://proceedings.neurips.cc/paper/2014/file/09c6c3783b4a70054da74f2538ed47c6-Paper.pdf}

\bibitem{DBLP:conf/bmvc/ParkWLK18}
Park, J., Woo, S., Lee, J., Kweon, I.S.: {BAM:} bottleneck attention module.
  In: British Machine Vision Conference 2018, {BMVC} 2018, Newcastle, UK,
  September 3-6, 2018. p.~147. {BMVA} Press (2018),
  \url{http://bmvc2018.org/contents/papers/0092.pdf}

\bibitem{NEURIPS2019_bdbca288}
Paszke, A., Gross, S., Massa, F., Lerer, A., Bradbury, J., Chanan, G., Killeen,
  T., Lin, Z., Gimelshein, N., Antiga, L., Desmaison, A., Kopf, A., Yang, E.,
  DeVito, Z., Raison, M., Tejani, A., Chilamkurthy, S., Steiner, B., Fang, L.,
  Bai, J., Chintala, S.: Pytorch: An imperative style, high-performance deep
  learning library. In: Wallach, H., Larochelle, H., Beygelzimer, A.,
  d\textquotesingle Alch\'{e}-Buc, F., Fox, E., Garnett, R. (eds.) Advances in
  Neural Information Processing Systems. vol.~32. Curran Associates, Inc.
  (2019),
  \url{https://proceedings.neurips.cc/paper/2019/file/bdbca288fee7f92f2bfa9f7012727740-Paper.pdf}

\bibitem{pinheiro2014recurrent}
Pinheiro, P., Collobert, R.: Recurrent convolutional neural networks for scene
  labeling. In: International conference on machine learning. pp. 82--90. PMLR
  (2014)

\bibitem{qin2021fcanet}
Qin, Z., Zhang, P., Wu, F., Li, X.: Fcanet: Frequency channel attention
  networks. In: Proceedings of the IEEE/CVF International Conference on
  Computer Vision. pp. 783--792 (2021)

\bibitem{recht2019imagenet}
Recht, B., Roelofs, R., Schmidt, L., Shankar, V.: Do imagenet classifiers
  generalize to imagenet? In: International Conference on Machine Learning. pp.
  5389--5400. PMLR (2019)

\bibitem{ridnik2021asymmetric}
Ridnik, T., Ben-Baruch, E., Zamir, N., Noy, A., Friedman, I., Protter, M.,
  Zelnik-Manor, L.: Asymmetric loss for multi-label classification. In:
  Proceedings of the IEEE/CVF International Conference on Computer Vision. pp.
  82--91 (2021)

\bibitem{roy2018recalibrating}
Roy, A.G., Navab, N., Wachinger, C.: Recalibrating fully convolutional networks
  with spatial and channel “squeeze and excitation” blocks. IEEE
  transactions on medical imaging  \textbf{38}(2),  540--549 (2018)

\bibitem{selvaraju2017grad}
Selvaraju, R.R., Cogswell, M., Das, A., Vedantam, R., Parikh, D., Batra, D.:
  Grad-cam: Visual explanations from deep networks via gradient-based
  localization. In: Proceedings of the IEEE international conference on
  computer vision. pp. 618--626 (2017)

\bibitem{touvron2021training}
Touvron, H., Cord, M., Douze, M., Massa, F., Sablayrolles, A., J{\'e}gou, H.:
  Training data-efficient image transformers \& distillation through attention.
  In: International Conference on Machine Learning. pp. 10347--10357. PMLR
  (2021)

\bibitem{NEURIPS2019_d03a857a}
Touvron, H., Vedaldi, A., Douze, M., Jegou, H.: Fixing the train-test
  resolution discrepancy. In: Wallach, H., Larochelle, H., Beygelzimer, A.,
  d\textquotesingle Alch\'{e}-Buc, F., Fox, E., Garnett, R. (eds.) Advances in
  Neural Information Processing Systems. vol.~32. Curran Associates, Inc.
  (2019),
  \url{https://proceedings.neurips.cc/paper/2019/file/d03a857a23b5285736c4d55e0bb067c8-Paper.pdf}

\bibitem{treisman1980feature}
Treisman, A.M., Gelade, G.: A feature-integration theory of attention.
  Cognitive psychology  \textbf{12}(1),  97--136 (1980)

\bibitem{vaswani2017attention}
Vaswani, A., Shazeer, N., Parmar, N., Uszkoreit, J., Jones, L., Gomez, A.N.,
  Kaiser, {\L}., Polosukhin, I.: Attention is all you need. Advances in neural
  information processing systems  \textbf{30} (2017)

\bibitem{wang2017residual}
Wang, F., Jiang, M., Qian, C., Yang, S., Li, C., Zhang, H., Wang, X., Tang, X.:
  Residual attention network for image classification. In: Proceedings of the
  IEEE conference on computer vision and pattern recognition. pp. 3156--3164
  (2017)

\bibitem{wang2020eca}
Wang, Q., Wu, B., Zhu, P., Li, P., Zuo, W., Hu, Q.: Eca-net: efficient channel
  attention for deep convolutional neural networks, 2020 ieee. In: CVF
  Conference on Computer Vision and Pattern Recognition (CVPR). IEEE (2020)

\bibitem{wang2021not}
Wang, Y., Huang, R., Song, S., Huang, Z., Huang, G.: Not all images are worth
  16x16 words: Dynamic transformers for efficient image recognition. Advances
  in Neural Information Processing Systems  \textbf{34} (2021)

\bibitem{WelinderEtal2010}
Welinder, P., Branson, S., Mita, T., Wah, C., Schroff, F., Belongie, S.,
  Perona, P.: {Caltech-UCSD Birds 200}. Tech. Rep. CNS-TR-2010-001, California
  Institute of Technology (2010)

\bibitem{woo2018cbam}
Woo, S., Park, J., Lee, J.Y., Kweon, I.S.: Cbam: Convolutional block attention
  module. In: Proceedings of the European conference on computer vision (ECCV).
  pp. 3--19 (2018)

\bibitem{xu2015show}
Xu, K., Ba, J., Kiros, R., Cho, K., Courville, A., Salakhudinov, R., Zemel, R.,
  Bengio, Y.: Show, attend and tell: Neural image caption generation with
  visual attention. In: International conference on machine learning. pp.
  2048--2057. PMLR (2015)

\bibitem{yang2019cross}
Yang, J., Ren, Z., Gan, C., Zhu, H., Lin, J., Parikh, D.: Cross-channel
  communication networks  (2019)

\bibitem{yang2016stacked}
Yang, Z., He, X., Gao, J., Deng, L., Smola, A.: Stacked attention networks for
  image question answering. In: Proceedings of the IEEE conference on computer
  vision and pattern recognition. pp. 21--29 (2016)

\bibitem{zamir2017feedback}
Zamir, A.R., Wu, T.L., Sun, L., Shen, W.B., Shi, B.E., Malik, J., Savarese, S.:
  Feedback networks. In: Proceedings of the IEEE conference on computer vision
  and pattern recognition. pp. 1308--1317 (2017)

\bibitem{zeiler2014visualizing}
Zeiler, M.D., Fergus, R.: Visualizing and understanding convolutional networks.
  In: European conference on computer vision. pp. 818--833. Springer (2014)

\bibitem{zhang2020putting}
Zhang, M., Tseng, C., Kreiman, G.: Putting visual object recognition in
  context. In: Proceedings of the IEEE/CVF Conference on Computer Vision and
  Pattern Recognition. pp. 12985--12994 (2020)

\bibitem{zhao2021recurrence}
Zhao, J., Fang, Y., Li, G.: Recurrence along depth: Deep convolutional neural
  networks with recurrent layer aggregation. Advances in Neural Information
  Processing Systems  \textbf{34},  10627--10640 (2021)

\end{thebibliography}
\end{document}